\documentclass[a4paper]{article}

\usepackage[english]{babel}
\usepackage{amsmath} 
\usepackage{graphicx}
\usepackage{fancyhdr}
\usepackage{setspace}
\usepackage{lineno}	 
\usepackage{lastpage}
\usepackage[margin=1in]{geometry}
\usepackage{caption}  
\usepackage[numbers,comma,sort&compress,super]{natbib}

{
	\fancyhead[L]{\myshorttitle}
	\fancyhead[R]{\small{\textit{\the\day.\the\month.\the\year}}}
	\fancyfoot[L]{\copyright \small{\textit{\myauthorshort}}}
	\fancyfoot[R]{Page \thepage\ of \pageref*{LastPage}}
	\cfoot{}
}

\fancypagestyle{firststyle}
{
	\fancyhead[L]{\copyright \small{\textit{\myauthorshort}}}
	\fancyhead[R]{Page \thepage\ of \pageref*{LastPage}}
	\fancyfoot[L]{\footnotesize \rule{0.25\textwidth}{0.4pt} \newline Corresponding author: $^{**}$A. M. Schweidtmann, E-mail: a.schweidtmann@tudelft.nl}
	\fancyfoot[C]{}
	\fancyfoot[R]{}
}

\renewenvironment{abstract}{\noindent\textbf{Abstract:}}{}

\newenvironment{acknowledgements}{\noindent\footnotesize\textbf{Acknowledgements}}{}

\usepackage[nohyperlinks]{acronym}
\acrodefplural{ANN}[ANNs]{artificial neural networks}
\acrodefplural{CNN}[CNNs]{convolutional neural networks}
\acrodefplural{GCN}[GCNs]{graph convolutional neural networks}
\acrodefplural{GNN}[GNNs]{graph neural networks}
\acrodefplural{MLP}[MLPs]{multi-layer perceptrons}
\acrodefplural{RNN}[RNNs]{Recurrent Neural Networks}
\usepackage{authblk}
\usepackage{polski}
\usepackage{dblfloatfix}
\usepackage{algorithm}
\usepackage{algpseudocode}
\usepackage[version=4]{mhchem}
\usepackage{siunitx}
\usepackage{subcaption}
\usepackage{booktabs}
\usepackage{tabulary}
\usepackage{supertabular}
\usepackage{eurosym}
\algdef{SE}{Begin}{End}{\textbf{begin}}{\textbf{end}}
\DeclareSIUnit\year{a}
\newcolumntype{P}[1]{>{\centering\arraybackslash}p{#1}}
\newcolumntype{M}[1]{>{\centering\arraybackslash}m{#1}}
\newcolumntype{N}[1]{>{\raggedright\arraybackslash}m{#1}}

\newcommand{\mytitle}{Flowsheet synthesis through hierarchical reinforcement learning and graph neural networks}
\newcommand{\myshorttitle}{Flowsheet synthesis through graph-based reinforcement learning}
\newcommand{\myauthor}{Laura Stops$^{1,*}$, Roel Leenhouts$^{1,*}$, Qinghe Gao$^1$, Artur M. Schweidtmann$^{1,**}$} 
\newcommand{\myauthorshort}{A. M. Schweidtmann}
\author{\myauthor}

\usepackage[colorlinks,linkcolor=blue,citecolor=blue,urlcolor=blue,pdftitle={\mytitle},pdfauthor={Artur M. Schweidtmann}]{hyperref} 
\usepackage{cleveref}

\begin{document}

\thispagestyle{firststyle}
	\begin{flushleft}\begin{large}\textbf{\mytitle}\end{large} \end{flushleft}
	\myauthor 
	
	\begin{flushleft}\begin{small}
			$^*$ contributed equally\\ [0.25cm]
			$^1$ Delft University of Technology, \\
			Department of Chemical Engineering, \\
			Van der Maasweg 9, \\
			Delft 2629 HZ, \\
			The Netherlands\\[0.25cm]
		\end{small}
	\end{flushleft}

\begin{abstract}
\noindent
Process synthesis experiences a disruptive transformation accelerated by digitization and artificial intelligence.
We propose a reinforcement learning algorithm for chemical process design based on a state-of-the-art actor-critic logic. 
Our proposed algorithm represents chemical processes as graphs and uses graph convolutional neural networks to learn from process graphs. 
In particular, the graph neural networks are implemented within the agent architecture to process the states and make decisions. 
Moreover, we implement a hierarchical and hybrid decision-making process to generate flowsheets, where unit operations are placed iteratively as discrete decisions and corresponding design variables are selected as continuous decisions. 
We demonstrate the potential of our method to design economically viable flowsheets in an illustrative case study comprising equilibrium reactions, azeotropic separation, and recycles.
The results show quick learning in discrete, continuous, and hybrid action spaces. 
Due to the flexible architecture of the proposed reinforcement learning agent, the method is predestined to include large action-state spaces and an interface to process simulators in future research. 
\end{abstract}

\noindent\textbf{Topical heading:} Process systems engineering
\newpage

\section{Introduction}
\label{sec:Introduction}
The chemical industry is approaching a disruptive transformation towards a more sustainable and circular future~\cite{Fantke.2021, Meramo.2021, Mohan.2021}. 
As a major contributor to global emissions, tremendous changes are required and the chemical industry needs to face a paradigm shift~\cite{Fantke.2021}.
This also requires rethinking regarding the conceptualization of novel processes~\cite{Martinez.2017, Meramo.2021}.
Simultaneously, innovations are pushed by new possibilities due to emerging digital technologies. 
Digitization and in particular \ac{AI} offer new possibilities for process design and therefore have the potential contribute to the transformation of chemical engineering~\cite{Schweidtmann.2021, Fantke.2021, Mohan.2021}. 

In the last decade, \ac{RL} has demonstrated its potential to solve complex decision-making problems, e.g., by showing human-like or even superhuman performance in a large variety of game applications~\cite{Mnih.2013, Kempka.2016, Silver.2018}.
\ac{RL} is a subcategory of \ac{ML} where an agent learns to interact with an environment based on trial-and-error~\cite{Sutton.2018}.
Especially since 2016, when DeepMind’s AlphaGo~\cite{Silver.2016} succeeded against a world-class player in the game Go, \ac{RL} has attracted great attention. 
In recent developments, \ac{RL} applications have proven to successfully compete with top-tier human players in even real-time strategy video games like StarCraft II~\cite{Vinyals.2019} and Dota 2~\cite{OpenAI.2019}.

The accomplishments of \ac{RL} in gaming have initiated significant developments in other research fields, including chemistry and chemical engineering.
In process systems engineering, \ac{RL} has been mainly applied to scheduling~\cite{Lee.2022, Hubbs.2020} and process control~\cite{Hoskins.1992, Spielberg.2017, Mowbray.2021, Sachio.2021a, Sachio.2021b}.
After first appearances of \ac{RL} for process control in the early 1990s~\cite{Hoskins.1992}, the development  was pushed with the rise of deep \ac{RL} in continuous control in games~\cite{Mnih.2015} and physical tasks~\cite{Lillicrap.2016}. 
Spielberg et al.~\cite{Spielberg.2017} first transferred deep \ac{RL} to chemical process control.
In recent works, the satisfaction of joint chance constraints~\cite{Mowbray.2021} and the integration of process control into process design tasks~\cite{Sachio.2021a, Sachio.2021b} \textit{via} \ac{RL} were considered.

In contrast to continuous process control tasks, \ac{RL} in molecule design is characterized by discrete decisions, such as adding or removing atoms. 
Several methods use \ac{RL} for the design of molecules with desired properties~\cite{You.2018, Popova.2018, Zhou.2019, Cao.2018, Olivecrona.2017}. 
First applications generate \ac{SMILES} strings using \ac{RL} agents with pre-trained neural networks~\cite{Olivecrona.2017, Popova.2018}. 
Zhou et al.~\cite{Zhou.2019} introduced a method solely based on \ac{RL}, thereby ensuring chemical validity. 
Recently, \ac{RL} based molecule design has been further enhanced in terms of exploration strategies~\cite{Pereira.2021} or by combining \ac{RL} with orientation simulations~\cite{Jeon.2020}. 
In another approach, You et al.~\cite{You.2018} introduced a \ac{GCPN} that represents molecules as graphs. 
It allows using \acp{GNN} to approximate the policy of the \ac{RL} agent and to learn directly on the molecular graph. 
Using \acp{GNN} on molecule graphs to predict molecule properties~\cite{Schweidtmann.2020, Gilmer.2017, Coley.2017, Yang.2019} has also shown promising results besides \ac{RL}. 
For example, Schweidtmann et al.~\cite{Schweidtmann.2020} achieved competitive results for fuel property prediction by concatenating the output of a \ac{GNN} into a molecule fingerprint and further passing it trough a \ac{MLP}.  

Graph representation and \ac{RL} are also applied in other engineering fields. For example, Ororbia and Warn~\cite{Ororbia_2021} represent design configurations of planar trusses as graphs in an \ac{RL} optimization task.

Recently, important first steps have been made towards using \ac{RL} to synthesize novel process flowsheets~\cite{Midgley.2020, Khan.2020, Goettl.2021b, Goettl.2021, Intemic.2022, Plathottam.2021}. 
Midgley~\cite{Midgley.2020} introduced the "Distillation Gym", an environment in which distillation trains for non-azeotropic mixtures are generated by a soft-actor-critic \ac{RL} agent and simulated in the open-source process simulator COCO.
The agent first decides whether to add a new distillation column to the intermediate flowsheet and subsequently selects continuous operating conditions. 
In an alternative approach to generate process flowsheets, Khan and Lapkin~\cite{Khan.2020} presented a value-based agent that chooses the next action by assessing its value, based on previous experience.
The agent operates within a hybrid action space, i.e., it makes discrete and continuous decisions.
In a recent publication, Khan and Lapkin~\cite{Khan.2022} introduced a hierarchical \ac{RL} approach to process design, capable of designing more advanced process flowsheets, also including recycles. 
A higher level agent constructs process sections by choosing sub-objectives of the process, such as maximizing the yield. 
Then, a lower level agent operates within these sections and chooses unit types and discretized parametric control variables that define unit conditions.
Due to the discretization, the agent operates only in a discrete action space.
As another approach to synthesize flowsheets with \ac{RL}, Göttl et al.~\cite{Goettl.2021b} developed a turn-based two-player-game environment called "SynGameZero". 
The interpretation of flowsheeting as a two-player game allowed them to reuse an established tree search \ac{RL} algorithm from DeepMind~\cite{Silver.2018}.
Recently, Göttl et al.~\cite{Goettl.2021} enhanced their work by allowing for recycles and utilizing \acp{CNN} for processing large flowsheet matrices.
Additionally, the company Intemic~\cite{Intemic.2022} has recently developed a "flowsheet copilot" that generates flowsheets iteratively, embededded in a 1-player-game. Intemic offers a web front-end in which raw materials and desired products can be specified.
Then, a \ac{RL} agents selects unit operations as discrete decisions using the economic value of the resulting process as objective. 
Furthermore, Plathottam et al.~\cite{Plathottam.2021} introduced a \ac{RL} agent that optimizes a solvent extraction process by selecting discrete and continuous design variables within predefined flowsheets.

One major gap in the previous literature on \ac{RL} for process synthesis is the state representation of flowsheets. 
We believe that a meaningful information representation is key to enable breakthroughs of \ac{AI} in chemical engineering~\cite{Schweidtmann.2021}. 
Previous works represent flowsheet in matrices comprising thermodynamic stream data, design specifications, and topological information~\cite{Goettl.2021}.
However, we know from computer science research that passing such matrices through \acp{CNN} is limited as they can only operate on fixed grid topologies, thereby exploiting spatial but not geometrical features~\cite{Bruna.2013}.
In contrast, \acp{GCN} handle differently sized and ordered neighborhoods~\cite{Zhou.2020} with the topology becoming a part of the network's input~\cite{Bronstein.2021}.
Since flowsheets are naturally represented as graphs with varying size and order of neighborhoods, \acp{GCN} can take their topological information into account.
Another gap in the literature concerns the combination of multiple unit operation types, recycle streams and a larger, hybrid action space.
While previous works proposed these promising techniques in individual contributions~\cite{Midgley.2020, Khan.2020,Khan.2022, Goettl.2021b, Goettl.2021, Plathottam.2021, Intemic.2022}, they have not yet been combined to a unified framework.

In this contribution, we represent flowsheets as graphs consisting of unit operations as nodes and streams as edges (c.f.\cite{Zhang.2018,Friedler.2019}). 
The developed agent architecture features a flowsheet fingerprint, which is learned by processing flowsheet graphs in \acp{GNN}.
Thereby, \ac{PPO}~\cite{Schulman.2017} is deployed with modifications to learn directly on graphs 
and to allow for hierarchical decisions.
In addition, we combine a hybrid action space, hierarchical actor-critic \ac{RL},  and graph generation in a unified framework.

\section{Reinforcement learning for process synthesis}
\label{sec:Methodology}
In this section, we introduce the methodology and the architecture of the proposed method. 
To apply \ac{RL} to process synthesis, the problem is first formulated as a \ac{MDP} which is defined by the tuple $M = \{ S, A, T , R \}$.
A \ac{MDP} consists of states $s \in S$, actions $a \in A$, a transition model $T: S \times A \rightarrow S$, and a reward function $R$~\cite{Sutton.2018}.
In the considered problem, states are represented by flowsheets graphs, while actions comprise discrete and continuous decisions. 
More specifically, the discrete decisions consist of selecting a new unit operation as well as the location where it is added to the intermediate flowsheet.
The continuous decisions are to define one or several specific continuous design variables per unit operation. 
For the environment, we implemented simple functions in Python to simulate the considered flowsheet.
Finally, a reward is calculated and returned to the agent. 

\begin{figure}
    \centering
    \includegraphics[width=0.6\textwidth]{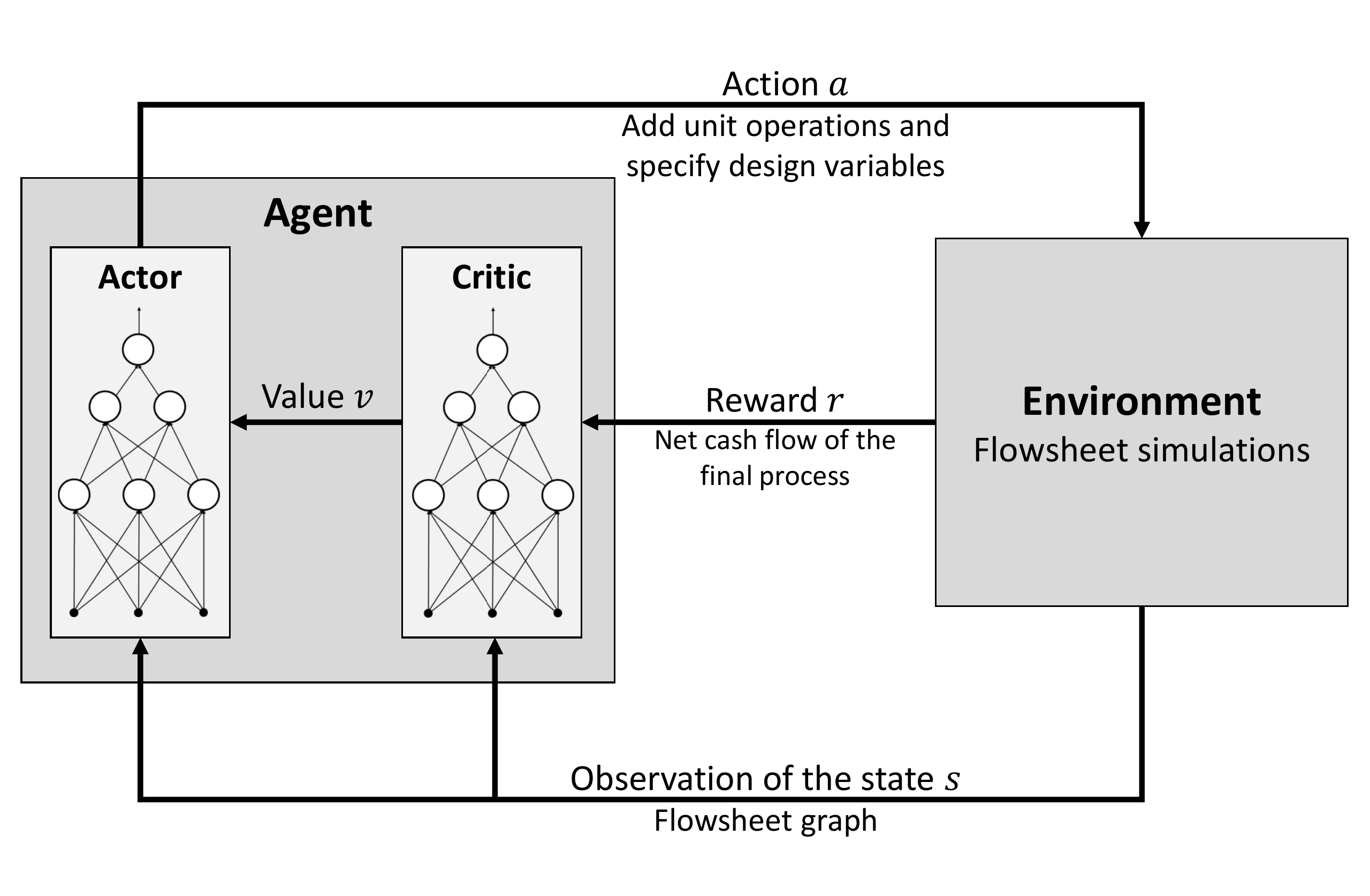}
    \caption{Agent-environment interaction in an actor-critic policy optimization approach for flowsheet synthesis.
    The agent approximates the policy and makes decisions. 
    Meanwhile, the critic estimates the value of the environment's state using the flowsheet graph, which is used to evaluate the agent's decisions. 
    Here, actor and critic both deploy graph convolutional neural networks.}
    \label{fig:actor-critic}
\end{figure}

While most \ac{RL} methods can be divided into value-based and policy-based approaches, actor-critic \ac{RL} takes advantage of both concepts~\cite{Sutton.2018}. 
In contrast to value-based \ac{RL} methods that cannot be easily adapted to continuous actions~\cite{Mnih.2016, Lillicrap.2016}, actor-critic approaches can learn policies for both, discrete and continuous action spaces and are thus also suitable for hybrid tasks~\cite{Fan.2019}.
Subsequently, several recent state-of-the-art policy optimization methods propose an actor-critic setup~\cite{Mnih.2016, Lillicrap.2016, Schulman.2017, Fujimoto.2018, Haarnoja.2018, Fan.2019}.
As shown in~\autoref{fig:actor-critic}, actor-critic agents consist of a critic that estimates the value function and an actor that decides for actions by approximating the policy~\cite{Sutton.2018}.

The \ac{RL} framework presented in this work is derived from the actor-critic \ac{PPO} algorithm by OpenAI~\cite{Schulman.2017}.
In \ac{PPO}, the objective function is clipped to prevent a collapse of the agent's performance during training.
To favor exploration, an entropy term~\cite{WILLIAMS.1991} is added to the loss function.
Additionally, the generalized estimation of the advantage $\hat{A}$~\cite{Schulman.2016} is used for updating the networks.

\subsection{State representation}
\label{sec:State}
The main feature of the proposed method is the representation of the states by directed flowsheet graphs. 
This characteristic allows us to process the states in \acp{GNN}, thereby taking topological information into account.

\begin{figure}
    \centering
    \includegraphics[width=0.4\textwidth]{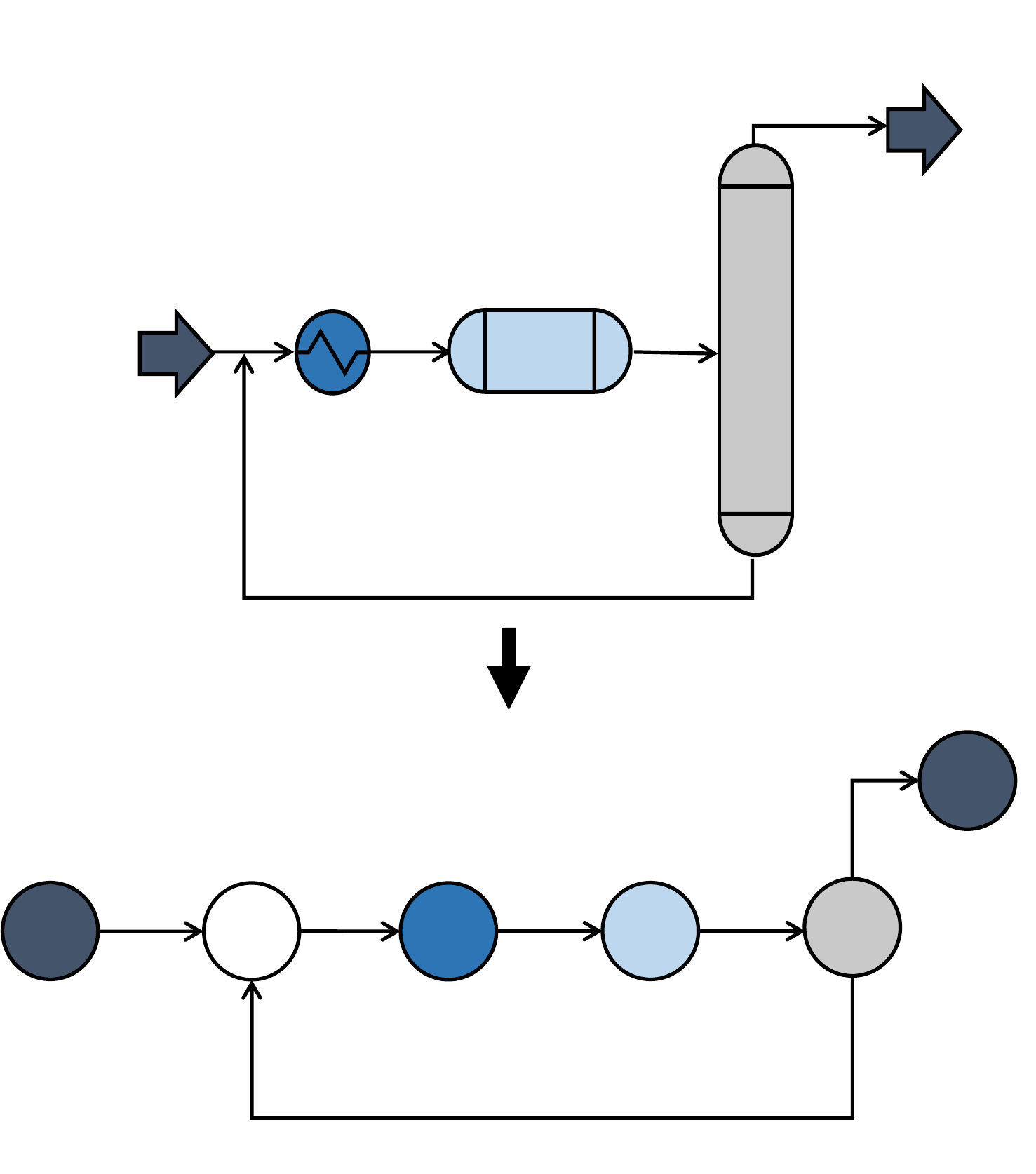}
    \caption{Example of a flowsheet displayed as a graph. Unit operations, feeds, and products are represented as nodes, whereas streams are represented as edges.}
    \label{fig:flowsheet-graph}
\end{figure}

\autoref{fig:flowsheet-graph} demonstrates the graph representation of flowsheets.
Feeds, products, and unit operations are represented by nodes, storing the type of unit operation and design variables. 
The edges include thermodynamic information about process streams, like temperature, molar flow, and molar fractions.

Intermediate flowsheets feature nodes of the type ``undefined``.
Whenever a new unit operation is added to the flowsheet, the resulting open streams are considered as such ``undefined`` nodes.
In subsequent steps, they represent possible locations for placing new unit operations.
Consequently, adding a new unit operation practically means replacing an ``undefined`` node with a defined one.

\subsection{Agent}
\label{sec:Agent}
At the heart of the proposed \ac{RL} method stands a hierarchical, hybrid actor-critic agent composed of multiple~\acp{GNN} and~\acp{MLP}. 
Its characteristics are introduced hereinafter.

\subsubsection{Hierarchical, hybrid action space}
\label{sec:ActionSpace}
\begin{figure}[!ht]
    \centering
    \includegraphics[width=0.5\textwidth]{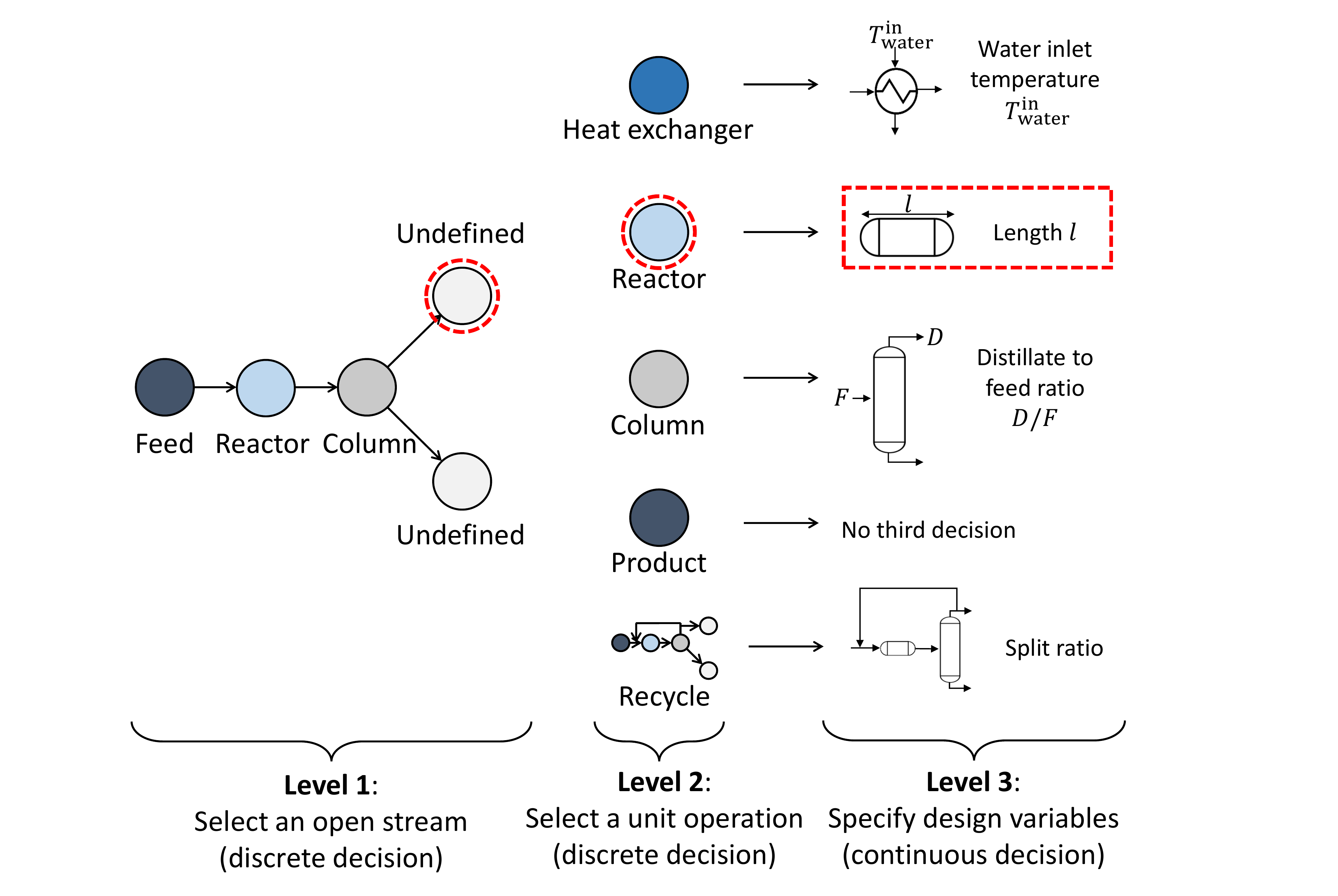}
    \caption{Hierarchical decision levels of the agent, starting from an intermediate flowsheet. In the first level, the agent selects a location where the flowsheet will be extended. Possible locations are open streams, represented by ``undefined`` nodes. In the presented flowsheet, both streams leaving the column can be chosen. Then, the agent selects a unit operation. Thereby, the options are to add a heat exchanger, a reactor, a  column, a recycle or to sell the stream as a product. Finally, a continuous design variable is selected for each unit operation. This third decision depends on which unit operation was selected previously.}
    \label{fig:decision_levels}
\end{figure}

The architecture of the agent is decisively affected by the considered hierarchical and hybrid action space.
The decision-making process is illustrated in~\autoref{fig:decision_levels}.
Every action consists of three levels of decisions:
(i) select a location, (ii) add a new unit operation, and (iii) define a continuous design variable.

In the first level, the agent decides for an open stream and thus for the location of the next flowsheet expansion.
As discussed in Section~\ref{sec:State}, open streams are identified by ``undefined`` nodes.
In the second level, the agent decides which type of unit operation will be added.
Thereby, the agent can choose to add a distillation column, a heat exchanger, or a reactor.
Furthermore, it can decide to add a recycle by introducing a splitter and a mixer into the flowsheet.
As a fifth option, the agent can declare the considered stream as a product.
If a unit operation is added, the third level decision is to specify the design variables of the corresponding unit operation. 
Although it is possible to set multiple design variables in this step, we chose to only set one variable for simplification reasons. 
Thus, one characteristic variable for each unit operation is defined in this step while all other variables are fixed. For the current implementation of the agent, the recycle stream is always inserted into the feed stream.
Whereas the first two levels are discrete decisions, the third level decisions are continuous. 
This combination of discrete and continuous decisions is referred to as hybrid action space. 

\subsubsection{Using GNNs to generate flowsheet fingerprints}
\label{sec:Fingerprint}
In \ac{RL}, every iteration of the agent-environment-interaction starts with the observation of the environment's state $s$, as shown in~\autoref{fig:actor-critic}.
In other approaches~\cite{Goettl.2021b, Goettl.2021, Midgley.2020, Khan.2022}, states or rather flowsheets are represented by vectors or matrices and, e.g., passed through \acp{CNN} for the observation step~\cite{Goettl.2021}.
Instead, in the herein presented approach, states are represented by flowsheet graphs (cf. Section~\ref{sec:State}).
To observe and process the therein stored information, the flowsheet graphs are passed through \acp{GCN} and encoded into a vector format called flowsheet fingerprint.
The advantage of using graphs and \acp{GCN} is that it allows operating in variable neighborhoods with different numbers and ordering of nodes, thereby taking spatial and spectral information into account~\cite{Bruna.2013, Zhou.2020, Bronstein.2021}.
Thus, we believe that graphs and \acp{GCN} are better suited for representing and processing the branched connectivity of flowsheets than passing matrices through \acp{CNN}. 

For this step, we transfer the method introduced by Schweidtmann et al.~\cite{Schweidtmann.2020}, who apply \acp{GNN} to generate molecule fingerprints, to flowsheets.
The approach utilizes the \ac{MPNN} proposed by Gilmer et al.~\cite{Gilmer.2017}.

\begin{figure*}[!ht]
    \centering
    \includegraphics[width=\textwidth]{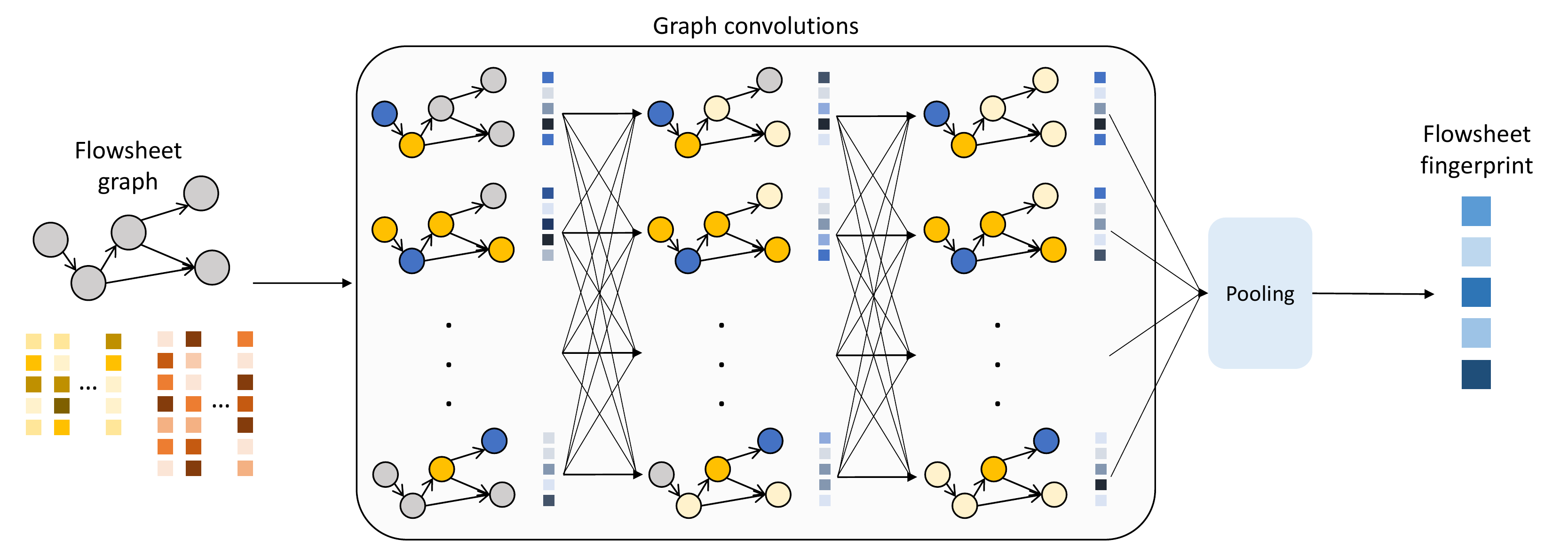}
    \caption{Flowsheet fingerprint generation derived from Schweidtmann et al.~\cite{Schweidtmann.2020}. The flowsheet graph is processed through an MPNN, using GCNs to perform message passing and update node embeddings. In the readout step, a pooling function is applied, resulting in a vector format, the flowsheet fingerprint.}
    \label{fig:fingerprint}
\end{figure*}

The overall scheme to process a flowsheet graph is displayed in~\autoref{fig:fingerprint} and consists of a message passing and a readout phase.
First, the flowsheet graph is processed through an \ac{MPNN}, using a \ac{GCN} with several layers to exchange messages and update node embeddings.
Afterward, a pooling function generates a vector format, the flowsheet fingerprint, in the readout phase. 
After several steps of message passing, sum-pooling is deployed for the subsequent readout phase.
Thereby, the node embeddings of the last layer are concatenated into a vector format, the flowsheet fingerprint.

\begin{figure*}
    \centering
    \includegraphics[width=\textwidth]{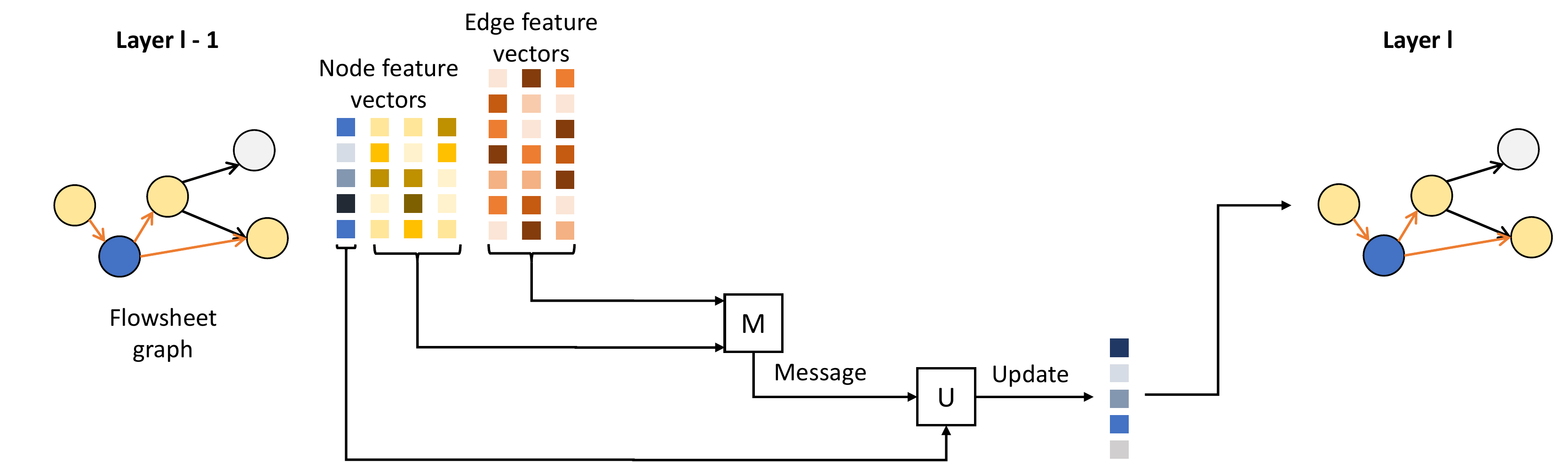}
    \caption{Update of the node embeddings during the message passing phase in a graph convolutional layer. The considered node is marked in blue and its neighbors in yellow. First, the information stored in the neighboring nodes and the respective edges is processed and combined through a message function M. Then, a message is generated to update the information embedded in the considered node through the update function U. The approach and its illustration follow a method proposed by Schweidtmann et al.~\cite{Schweidtmann.2020}.}
    \label{fig:gcn}
\end{figure*}

For every step in the message passing phase, first the node and edge features of the neighborhood of each node in the flowsheet graph are processed.
Therefore, \acp{GCN} are utilized to exchange and update information in the message passing phase.
The functionality of a graph convolutional layer is illustrated in \autoref{fig:gcn}, following Schweidtmann et al.~\cite{Schweidtmann.2020}.
The figure visualizes the procedure to update the node embeddings of the blue node. Therefore, the information stored in the yellow neighboring nodes and the corresponding edges is processed and combined to a message through the message function M. 
Then, the considered node is updated through the message in the update function U.
In each layer of a \ac{GCN}, this procedure is conducted for every node of the graph.

\subsubsection{Hierarchical agent architecture}
\label{sec:Architecture}
For the architecture of the agent, a structure suggested by Fan et al.~\cite{Fan.2019} for hierarchical and hybrid action spaces is used.
Thereby, individual \acp{MLP} are applied for each level of decisions and one \ac{MLP} is applied as a critic to evaluate the decisions.

\begin{figure*}[!ht]
    \centering
    \includegraphics[width=\textwidth]{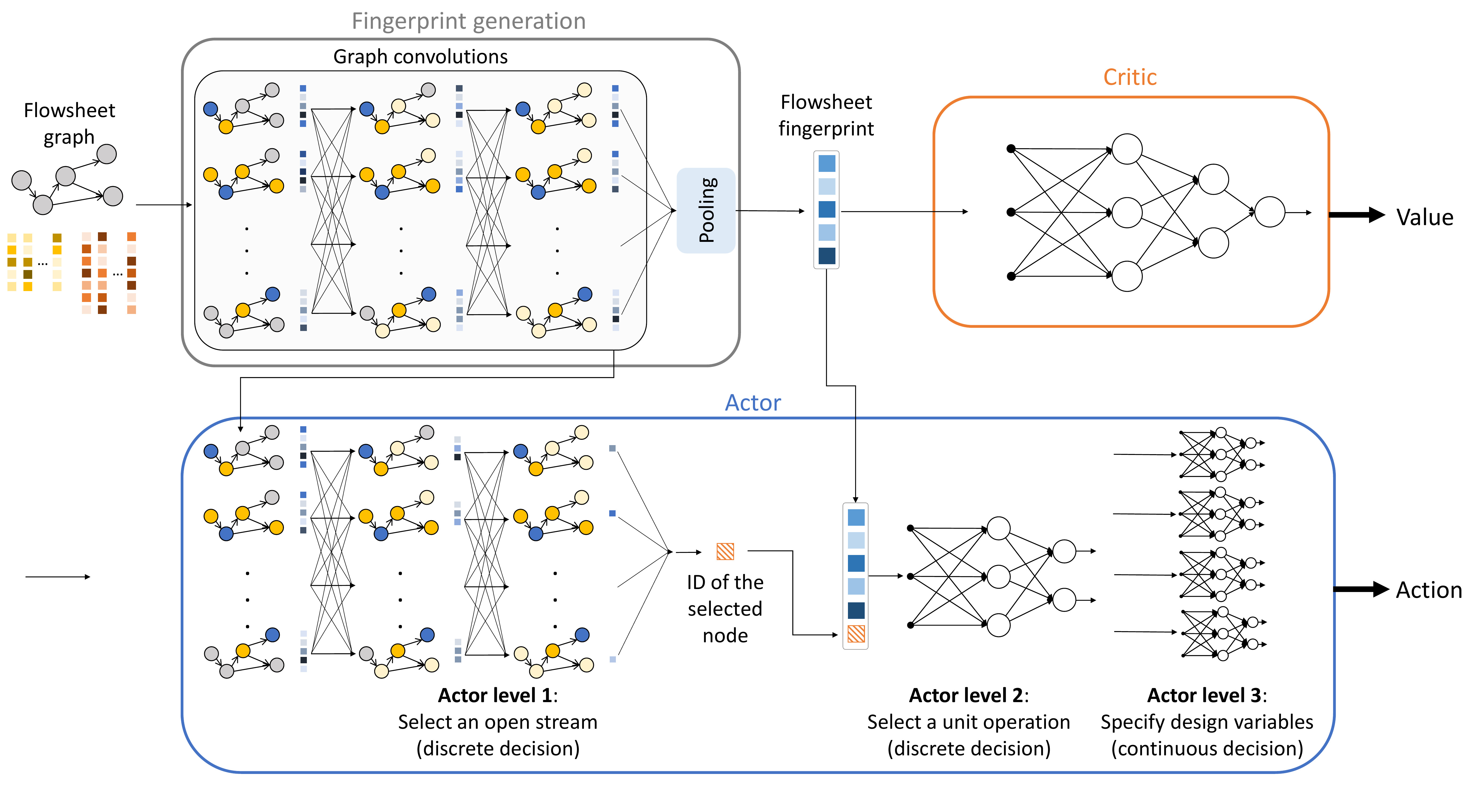}
    \caption{Architecture of the deployed actor-critic agent. First, a GNN is used to process the graph representation of the flowsheet into a flowsheet fingerprint. While the critic estimates the value of the fingerprint in one linear MLP, the actor takes three levels of decisions. The first decision is to choose a location for expanding the flowsheet. Practically, this means selecting the ID of a node representing an open stream. The selected node ID is combined with the flowsheet fingerprint and passed through an MLP for the second level decision of choosing a type of unit operation. Finally, a continuous design variable of the unit is chosen. Thereby, a different MLP is used for each unit type.}
    \label{fig:agent}
\end{figure*}

The architecture of the actor-critic approach is illustrated in~\autoref{fig:agent}.
In the ``fingerprint generation'' step, the state represented by a flowsheet graph is processed to a flowsheet fingerprint through a \ac{GCN}~(cf. Section~\ref{sec:Fingerprint}).
Additionally, the updated graph resulting from the message passing phase of the fingerprint generation is passed to the ``actor'' step. 
Therein, the updated graph is further processed by an additional \ac{GCN}. 
This represents the first level of the actor which is to select an open stream to further extend the flowsheet.
Thereby, the method takes advantage of the graph representation in which open streams end in ``undefined`` nodes.
In the \ac{GCN} of the first level decision, the number of node features is reduced to one (cf. related literature on node classification tasks~\cite{Zhou.2020}).
Furthermore, all nodes which do not correspond to open streams are filtered out. 
The remaining node feature of each nodes in the last \ac{GCN} layer represents its probability to be chosen as the location for adding a new unit.
Then, the ID of the selected node is concatenated with the previously computed flowsheet fingerprint before it is passed on to the second and third level actors as input.

The second level actor consists of a \ac{MLP} that returns probabilities for each unit operation to be chosen. 
For each type of unit operation, an individual \ac{MLP} is set up as the actor for the third level decision.
Thereby, the third level \acp{MLP} take the concatenated vector including the flowsheet fingerprint and the ID of the selected location as an input.
They return two outputs which are interpreted as parameters, $\alpha$ and $\beta$, describing a beta distribution $B\left(\alpha,\beta\right)$~\cite{Gupta.2004}.
Based on this distribution, a continuous decision regarding the respective design variable is made.

The critic that estimates the value of the original state is displayed in the upper half of~\autoref{fig:agent}. 
Therefore, the flowsheet fingerprint is passed through another \ac{MLP}. 
This value is an estimation of how much reward is expected to be received by the agent until the end of an episode when starting at the considered state and further following the current policy~\cite{Sutton.2018}.
In our approach, we utilize the value to compute the generalized advantage estimation $\hat{A}$ introduced by Schulman et al.~\cite{Schulman.2016}.
It tells whether an action performed better or worse than expected and is used to calculate losses of the actor's networks.
By comparing the value to the actual rewards, an additional loss is computed for the critic.

\subsection{Agent-environment interaction}
\label{sec:Implementation}
The interaction between the environment and the hierarchical actor-critic agent is further clarified in Algorithm~\ref{pseudocode:interaction}. After the environment is initialized with a feed, the flowsheet is generated in an iterative scheme. The agent first observes the current state $s$ of the environment and chooses actions $a$ for all three hierarchical decision levels by sampling. The agent returns the probabilities and the selected actions as well as the value $v$ of the state.
\begin{algorithm}[t]
    \caption{Pseudocode of the agent-environment interaction.}\label{pseudocode:interaction}
    \begin{algorithmic}
        \State $done$ = False
        \While {not $done$}
        \State observe state $s$
        \State actions $a$, probs $p$, value $v$ = \Call{Agent}{$s$}
        \State new state $s'$, reward $r$, $done$ = \Call{Env}{$a$}
        \State store transition ($s, a, p, r, done$) in memory
        \EndWhile
        \Statex
        \Function{Agent}{state $s$}
        \For{level=1,2,3}
        \State probs $p_{\mathrm{level}}$ = actor($s$)
        \State action $a_{\mathrm{level}}$ = sample($p_{\mathrm{level}}$)
        \EndFor
        \State value $v$ = critic($s$)
        \State \Return $a$, $p$, $v$
        \EndFunction
        \Statex
        \Function{Env}{actions $a$}
        \State next state $s'$ = SimulateFlowsheet($a$)
        \If{no more open streams}
        \State $done$ = True
        \State reward $r$ = NetCashFlow($s'$)
        \If{reward $r$ $<$ 0}
        \State reward $r$ = reward $r$ / 10
        \EndIf
        \Else
        \State reward $r$ = 0 \euro \textit{\footnotesize}
        \EndIf
        \State \Return $s'$, $r$,  $done$
        \EndFunction
    \end{algorithmic} 
\end{algorithm}

In the next step, the actions are applied to the environment.
Therefore, the next state $s'$ is computed by simulating the extended flowsheet.
Additionally, the environment checks whether any open stream is left in the flowsheet, indicating that the episode is still to be completed.
Since the weights of the agent's networks are randomly initialized, early training episodes can result in very large flowsheets.
Thus, the total number of units is limited to 25 as additional guidance. 
If a flowsheet exceeds this number, all open streams are declared as products.

Additionally, the environment calculates the reward that depends on whether the flowsheet is completed or not.
If the net cash flow is positive, the reward equals the net cash flow. 
If the net cash flow is negative, the reward equals the net cash flow divided by a factor 10.
This procedure is implemented in order to encourage exploration of the agent. 
For the intermediate steps during the synthesis, process rewards of zero are given to the agent.
After each iteration, the transition is stored in a batch and later used for batch learning. 

\subsection{Training}
\label{sec:Training}
The presented method, including the flowsheet simulations, is implemented in Python 3.9. 
The training procedure is adapted from \ac{PPO} by OpenAI~\cite{Schulman.2017}.
It consists of multiple epochs of minibatch updates, whereby the minibatches result from sampling on the transition tuples stored in the memory. 
The agent's networks are thereby updated by gradient descent, using a loss function derived from summing up and weighting all losses of the individual actors, their entropies, and the loss of the critic.

\section{Case study}  
\label{sec:Process}
The proposed method is demonstrated in an illustrative case study considering the production of \ac{MeOAc}, a low-boiling liquid often used as a solvent~\cite{Methylacetate}.
In an industrial setting, \ac{MeOAc} is primarily produced in reactive columns by esterification of \ac{HOAc}~\cite{Huss.2003, Agreda.1982}.
For illustration, we consider only simplified flowsheets that use separate units for reaction and separation. 
\subsection{Process simulation}
\label{sec:Env}
For computing new states and rewards, the flowsheets generated by the agent are simulated in Python.
Therefore, we implemented a model for each type of unit operation that can be selected in the second level decision. 
In our case study, the agent can decide to place reactors, distillation columns, and heat exchangers.
Furthermore, the agent can add recycles or sell open streams as products.
\paragraph{Reactor.}
\label{sec:Reactor}
The reactor is modeled as a \ac{PFR}, in which the reversible equilibrium reaction shown in~\autoref{eq:reaction} takes place. 
\begin{equation}
    \ce{HOAc + MeOH <=> MeOAc + H2O}
    \label{eq:reaction}
\end{equation}
\ac{MeOAc} and its by-product \ac{H2O} are produced by esterification of \ac{HOAc} with \ac{MeOH} under the presence of a strong acid.
To calculate the composition of the process stream leaving the \ac{PFR}, we formulated a boundary value problem, depending on the reaction rate, and manually implemented a fourth-order Runge-Kutta method with fixed step-size as solver.
Thereby, the reactor is modeled isothermal, based on the temperature of the inflowing stream. The reaction kinetics are based on Xu and Chuang~\cite{Xu.1996}.

The length of the \ac{PFR} is specified by the agent as the continuous third level decision within the range of \SIrange{0.05}{20}{\meter}.
Thereby, the relation of the cross-sectional area $A$ of the \ac{PFR} to the molar flow $\dot{N}$ passing through it is fixed to $A/\dot{N} = \SI{0.1}{\square\meter\second\per\mole}$.
Notably, the length of the reactor significantly influences the conversion in the \ac{PFR}.
In addition, the equilibrium of the considered reaction depends on the temperature of the process stream which thus affects the reaction rate and the conversion in the \ac{PFR}.
Thereby, the temperature of the process stream can be influenced by heat exchangers upstream of the reactor.

\paragraph{Heat exchanger.}
\label{sec:Heater}
In the heat exchanger, heat is transferred between the process stream and a water stream.
The continuous third level decision specifies the inlet temperature of the water and thus also whether the process stream is cooled or heated.
To avoid evaporation of the process stream, the inlet water temperature is  chosen within the range of \SIrange{5}{53.8}{\celsius}, where the upper limit corresponds to the lowest possible boiling point of the considered quarternary system.
The heat exchanger model computes the heat duty, the required heat transfer area, and the outlet temperature of the process stream. The model is based on a countercurrent flow, shell and tube heat exchanger \cite{smith2016chemical}. 
A typical heat transfer coefficient of \SI{568}{\watt\per\kelvin\square\meter} is used \cite{seider.2016}.
Additionally, we assume that the process stream always approaches the water stream temperature within \SI{5}{\kelvin} in the heat exchanger.

\begin{figure}
    \centering
    \includegraphics[width=0.4\textwidth]{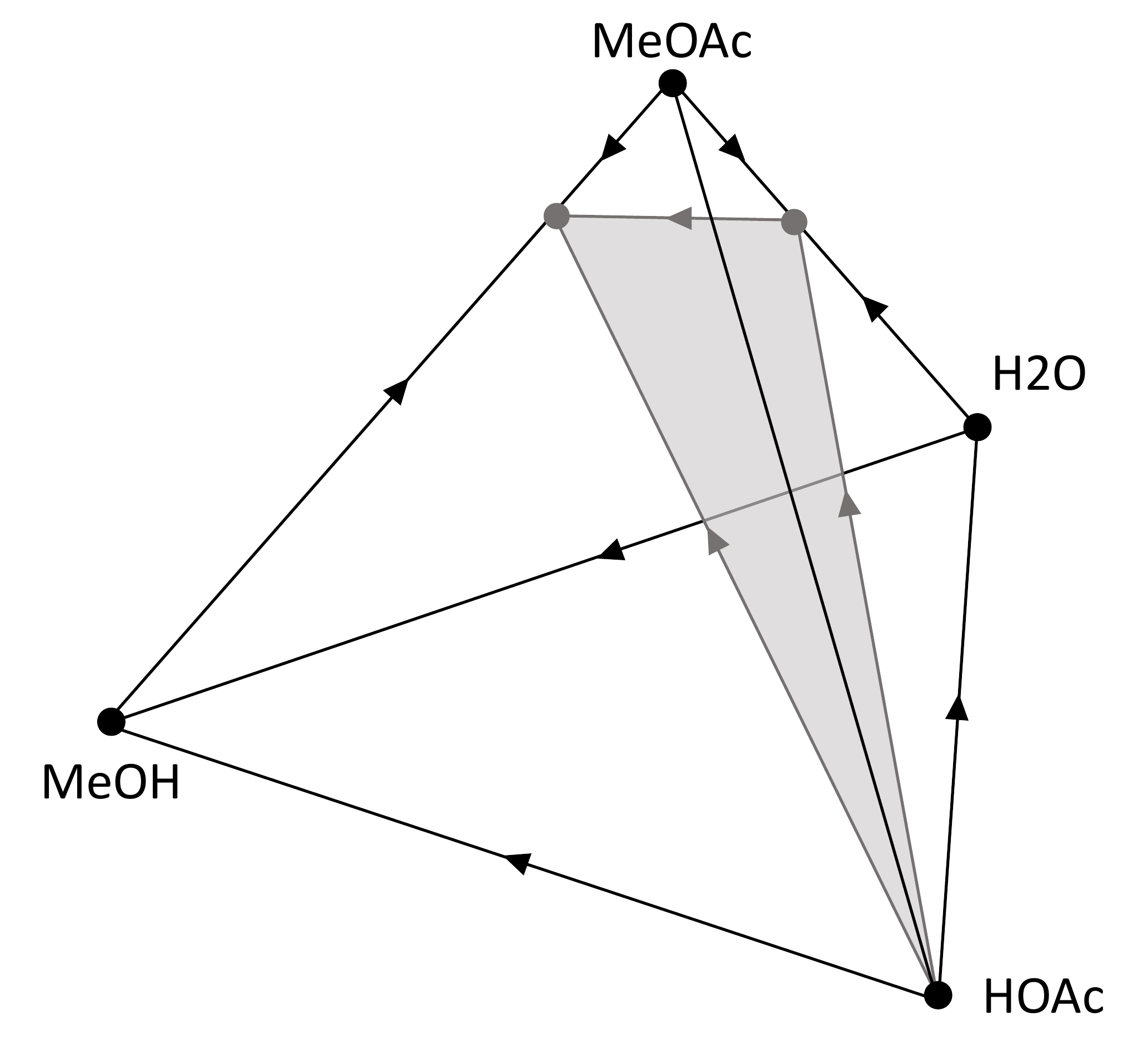}
    \caption{Vapor-liquid-equilibrium in the quarternary system consisting of \ac{MeOAc}, \ac{HOAc}, \ac{H2O} and \ac{MeOH} at \SI{1}{\bar}. The gray surface markes the distillation boundary spanned by the two azeotropic points and the fourth component \ac{HOAc}, spliting the diagram into two distillation regimes.}
    \label{fig:distillation-regime}
\end{figure}

\paragraph{Distillation column.}
\label{sec:Column}
The distillation column is deployed to separate the quarternary system \ac{MeOAc}, \ac{MeOH}, \ac{HOAc}, and \ac{H2O}.
The vapor-liquid equilibrium of the system is displayed in~\autoref{fig:distillation-regime}.
It contains two binary minimum azeotropes between  \ac{MeOAc} and \ac{H2O}, and respectively between \ac{MeOAc} and \ac{MeOH}. 
As shown in~\autoref{fig:distillation-regime}, the azeotropes split up the separation task into two distillation regimes.
To simplify the problem, we follow the assumption made by Göttl et al.~\cite{Goettl.2021} that the distillation boundary can be approximated by the simplex spanned between both azeotropes and the fourth component, \ac{HOAc}.

We implemented a shortcut column model using the $\infty / \infty$ analysis~\cite{Bekiaris.1993, Ryll.2012, Burger.2013}.
The only remaining degree of freedom in the  $\infty / \infty$ model is the  distillate to feed ratio $D/F$.
It is set by the agent in the continuous third level decision within a range of 0.05 to 0.95.

\paragraph{Recycle.}
\label{sec:Recycle}

The agent can also select to recycle an open process stream back to the feed stream.
Thereby, the ratio of the considered stream that will be recycled is selected by the agent in the third level decision.
The recycle is modelled by adding a splitting unit and a mixing unit to the flowsheet.
First, the considered stream is split up in a recycle stream and a purge stream. 
The latter one ends in a new "undefined" node. 
To simulate the recycle, a tear stream is initialized.
Then, the Wegstein method~\cite{wegstein.1958} is used to solve the recycle stream flow rate iteratively. 
When the Wegstein method is converged, the tear stream is closed and the recycle stream is fed into the feed stream by the mixing unit.
This method is based on the implementation of flexsolve \cite{YoelCortes-Pena2019Flexsolve:Solvers.}. 

\subsection{Reward}
\label{sec:Reward}
The reward assesses the economic viability of the generated process, following Seider et al.~\cite{seider.2016} for calculating annualized cost and Smith~\cite{smith2016chemical} for estimating unit capital costs.
After completing a flowsheet by specifying all open streams as products, the agent receives a final reward. 
This final reward $r$ represents an approximate net cash flow of the process within one year. If this net cash flow is negative, it is reduced by a factor 10 to encourage exploration of the agent.
The economic value of incomplete flowsheets is more difficult to estimate because it may depend on future actions. 
Thus, a reward of zero is given after every single action since the actual value of an action can only be assessed when an episode is complete.
As shown in~\autoref{eq:reward}, the final reward includes costs for units and feeds as well as revenue for sold products.
\begin{equation}
    r = \sum P_{\mathrm{\scriptscriptstyle products}} - \sum C_{\mathrm{\scriptscriptstyle feed}} - \sum\left(U+0.15*I\right)_{\mathrm{\scriptscriptstyle units}}
    \label{eq:reward}
\end{equation}
The values of the products are estimated by an s-shaped price function $P$, depending on the purity of the considered streams. The pure component price $C$ is used to compute the cost of the raw material stream.
The annualized cost is computed by adding the annual utility costs $U$ and the total capital investment $I$ multiplied by a factor 0.15 \cite{seider.2016}. 
Furthermore, the reward is used to teach the agent to make feasible decisions.
Whenever infeasible actions are selected that cause the simulation to fail, e.g., if the reactor simulation fails due to bad initial values in the solver, the episode is interrupted immediately and a  negative reward of~\num{-10}~Mio~\euro~is given. 
When the agent decides to not add units at all and just sell the feed streams, the same penalty is given to prevent the agent from falling into this trivial local optimum.

Notably, the considered case study is meant to facilitate illustration and the considered parameter values for prices are only approximations. 

\section{Results \& discussion}
\label{sec:Results}
In this section, we present and analyze the learning behavior of the developed agent. 
For investigating  all single parts of the agent, the training procedure was first conducted in a discrete action space, consisting of the first and second hierarchical decision levels.
Afterward, the same procedure was conducted in a continuous action space which only includes the third decision level.
Finally, all decision levels are combined to the hybrid action space.
In all runs, the environment was initialized with a feed consisting of an equimolar binary mixture of \ac{MeOH} and \ac{HOAc}. 
The feed's molar flow rate was set to \SI{100}{\mole\per\second} and its temperature to \SI{27}{\celsius}.

The proposed learning process and the agent architecture include several hyperparameters that are listed in the appendix in~\autoref{tab:hyperparameters}.
The selected hyperparameters are based on literature~\cite{Schweidtmann.2020, Gilmer.2017, Schulman.2017, Medium.2018}.

\subsection{Flowsheet generation in a discrete action space}
\label{sec:ResultsDiscrete}
To investigate the agent's behavior in a discrete action space, the third level actor was deactivated and only the first and second level decisions were conducted.
Thus, in each step, the agent selected a location for a new unit operation as well as its type.
Thereby, fixed values for the unit's continuous design variables were used. 
They are displayed in~\autoref{tab:fixed_parameters}.
\begin{table*}
    \centering
    \caption{Fixed continuous design variables for each unit type during the training in a discrete action space. This selection replaces the third level decision.}
    \begin{tabular}{llccc}
    \toprule
     Unit operation  & Design variable & Symbol & Unit & Fixed value \\
     \midrule
      Heat exchanger  & Water inlet temperature & $T^{in}_{\mathrm{water}}$ & \si{\celsius} & 32\\
      Reactor  & Reactor length & $l$ & \si{\meter} & 10\\
      Column   & Distillate to feed ratio & $D/F$ & - & 0.5\\
      Recycle & Recycling ratio & - & - & 0.9\\
      \bottomrule
    \end{tabular}
    \label{tab:fixed_parameters}
\end{table*}
Throughout the presented case study, constant pressure of~\SI{1}{\bar} was assumed.
The agent was trained in \num{10000} episodes with the procedure described in~\ref{sec:Implementation}.

\begin{figure}
    \centering
    \includegraphics[width=0.7\textwidth]{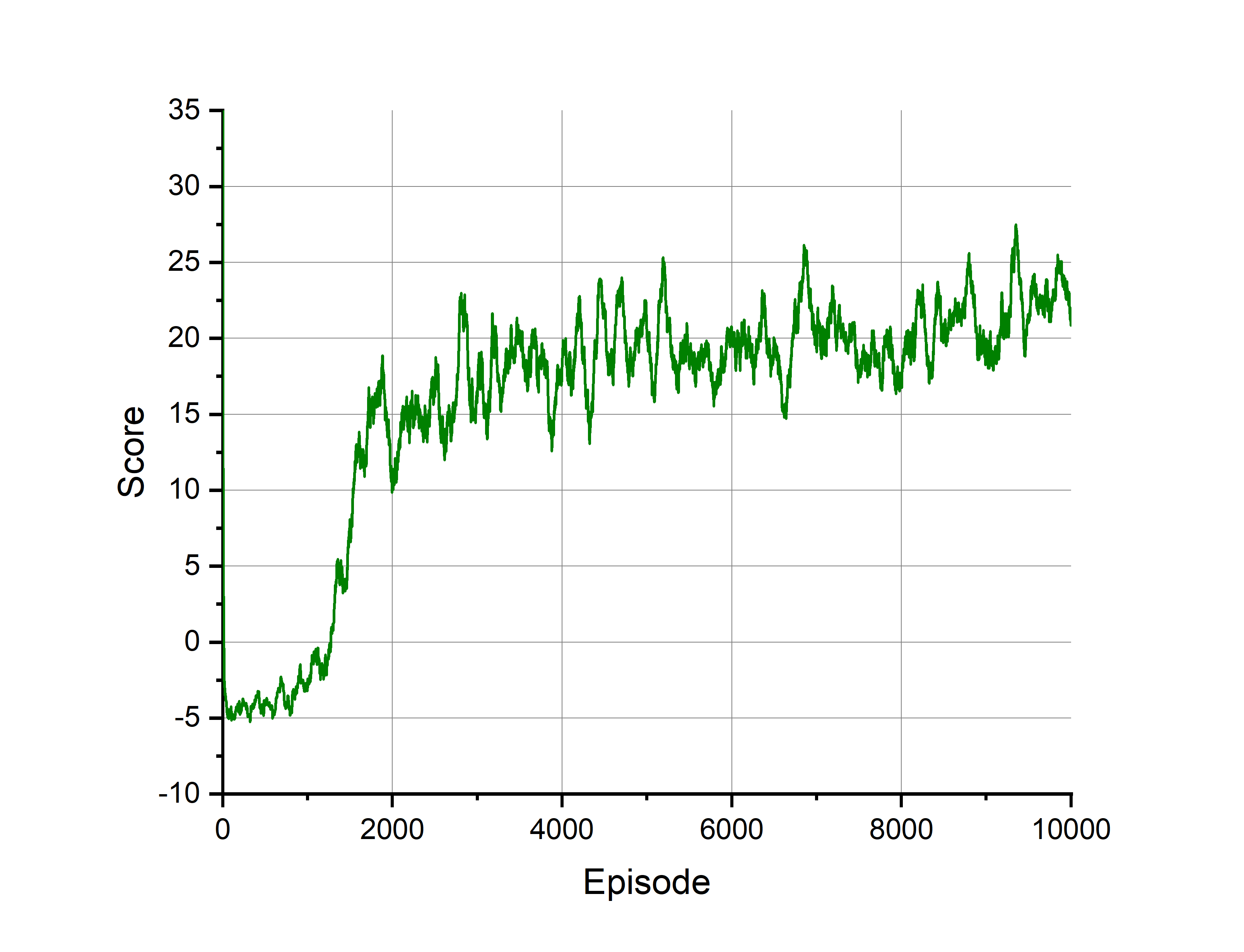}
    \caption{Learning curve of the agent in a discrete action space over \num{10000} episodes. It shows the scores of the generated flowsheets, averaged over 50 episodes. The score of each episode corresponds to the reward which is the estimated net cash flow. An episode is a sequence of actions to generate a flowsheet, starting with a feed.}
    \label{fig:lc_discrete}
\end{figure}

\autoref{fig:lc_discrete} shows the learning curve of the agent in the discrete action space. 
The displayed scores correspond to the reward which is the estimated net cash flow of the final process.
Thus, they are a measure of the economic viability of the final process.

During the first \num{2000} episodes, the learning curve rises almost exponentially. 
In this early training stage, the agent produces predominantly long flowsheets and often reaches the maximum allowed number of unit operations.
However, throughout the training the agent learns that shorter flowsheets are economically more valuable.
Soon, the agent mainly produces flowsheets with a positive score, meaning that the final process is economically viable.
Afterward, the learning curve still rises but only in minor scales.
One reason for the marginal improvements could be that the agent mainly exploits its experience at this time while still finding slightly better flowsheets through exploration.

\begin{figure} 
    \centering
        \includegraphics[width=0.4\textwidth]{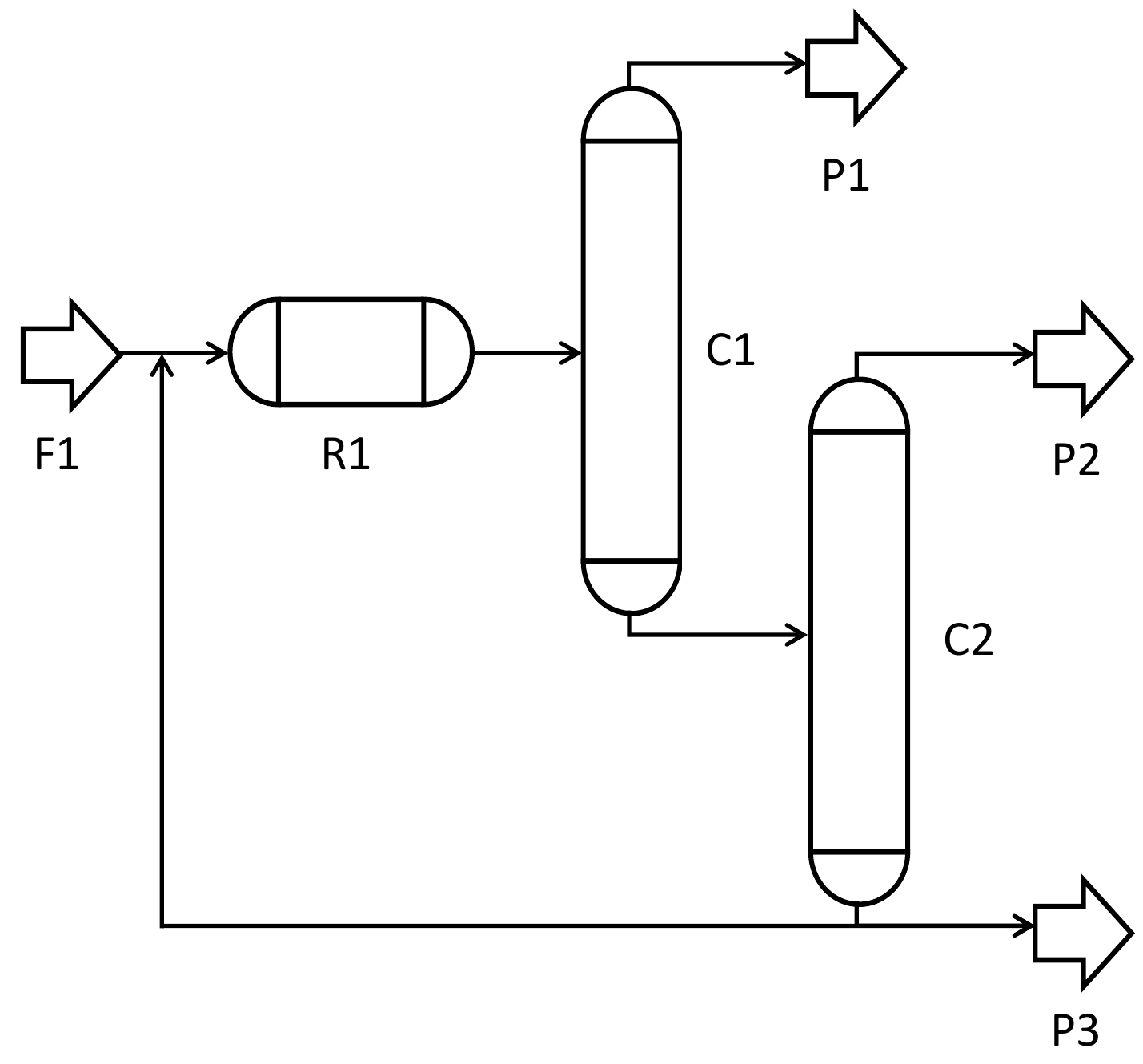}
        \caption{Best Flowsheet generated by the agent in a discrete action space after training for \num{10000} episodes. In a reactor (R1), \ac{MeOAc} and its side product \ac{H2O} are produced from the feed (F1). Then, the resulting quarternary mixture is split up in two columns (C1 and C2). Parts of the third product stream (P3) are recycled and mixed with the feed stream.}
    \label{fig:flowsheets_discrete}
\end{figure}

The best flowsheet the agent generated throughout training is displayed in~\autoref{fig:flowsheets_discrete}. 
The depicted process first uses a reactor (R1) to produce \ac{MeOAc} and its side product \ac{H2O} from the feed (F1).
Then, the resulting quarternary mixture is split up in two distillation columns.
The distillate (P1) of the first column (C1) is enriched with \ac{MeOAc} but also includes \ac{MeOH} and \ac{H2O}.
The bottom product of the first column is further split up in a second column (C2) to produce a mixture of \ac{H2O} and \ac{MeOH} in the distillate (P2) and pure \ac{MeOH} in the third product stream (P3). 
90 \% of the latter product is recycled and mixed with the feed stream.
During the training, the agent learned, for example, that heat exchangers do not add value to the flowsheet.

\subsection{Flowsheet generation in a continuous action space}
\label{sec:ResultsConti}
\begin{figure}
    \centering
    \includegraphics[width=0.4\textwidth]{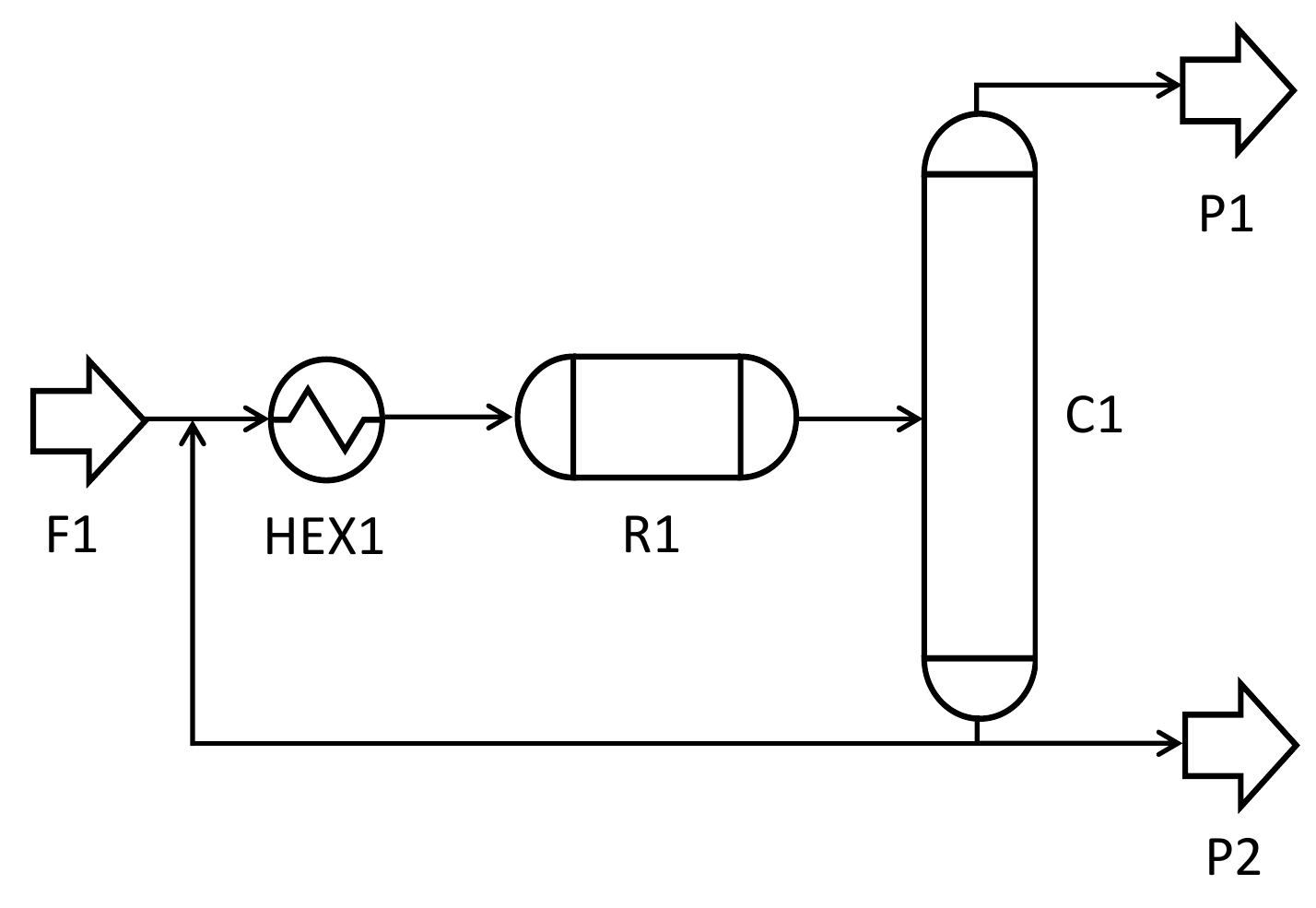}
    \caption{Fixed flowsheet structure during the training in a continuous action space. It consists of a heat exchanger (HEX1), a reactor (R1) and a column (C1). The bottom product (P2) is split up and partially recycled.}
    \label{fig:flowsheet_conti}
\end{figure}

\begin{figure}
    \centering
    \includegraphics[width=0.7\textwidth]{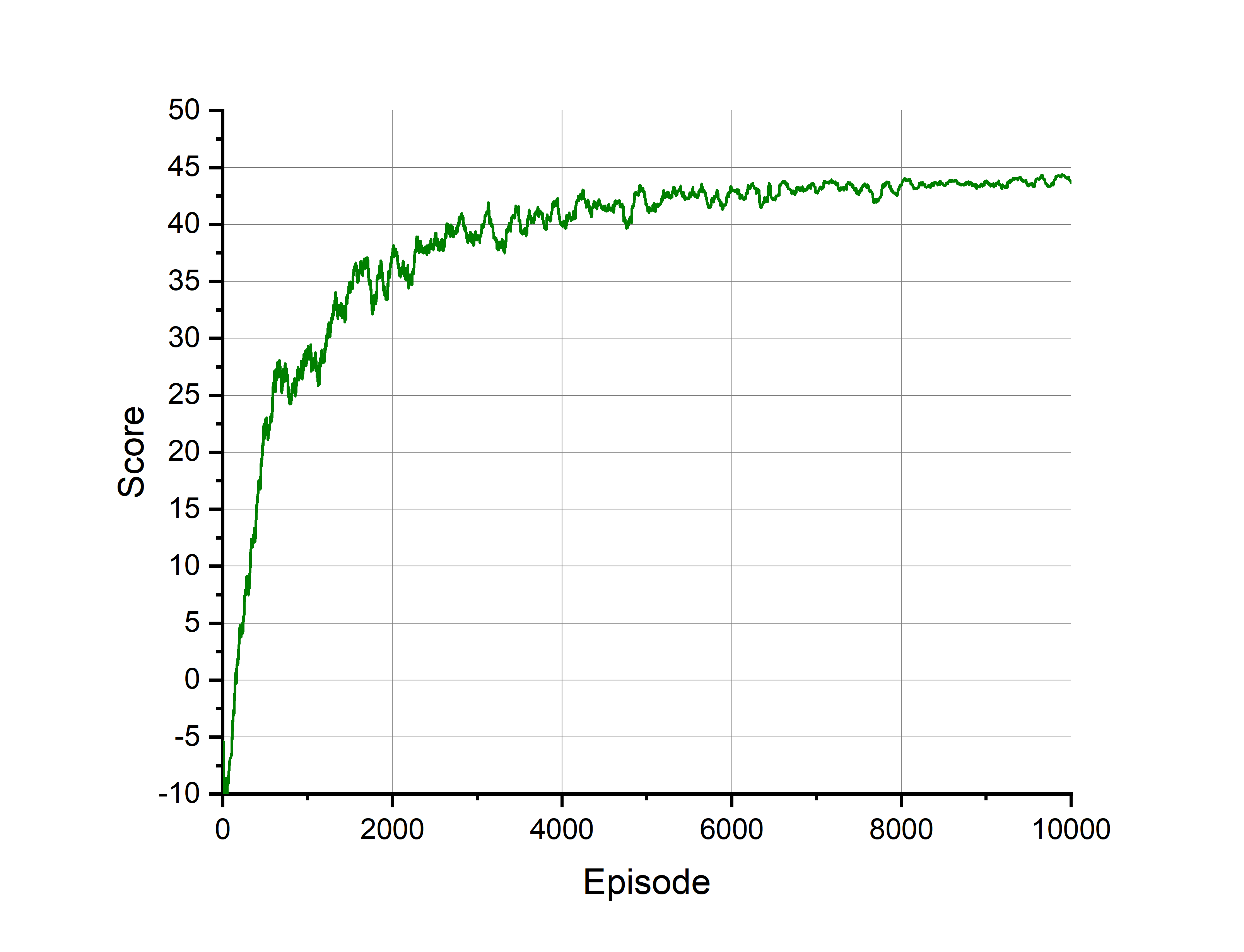}
    \caption{Learning curve of the agent in a continuous action space over \num{10000} episodes. Analogously to~\autoref{fig:lc_discrete}, it shows the scores of the generated flowsheets, averaged over \num{50} episodes.}
    \label{fig:lc_conti}
\end{figure}

The third level actor was investigated by deactivating the first and second level actors and thus only including continuous decisions. 
Therefore, the sequence of unit operations in the flowsheet was fixed, as shown in~\autoref{fig:flowsheet_conti}, and only the continuous design variables defining each unit were selected by the agent.
Within this structure, the agent was trained for \num{10000} episodes. 
Similar to the findings in the discrete action space, the agent learns quickly at the beginning of the training.
After the steep increase, the policy starts to converge and is almost constant after \num{10000} episodes. 
The resulting learning curve of the continuous agent is displayed in~\autoref{fig:lc_conti}, showing the scores of the final flowsheets averaged over \num{50} episodes.

\autoref{tab:parameters_conti} lists the continuous design variables of the best flowsheet the agent observed throughout the training. In the heat exchanger~(HEX1), the feed is slightly heated before entering the reactor.
With a length of \SI{5.24}{\meter}, the reactor~(R1) is relatively short compared to the allowed length range of \SIrange{0.05}{20}{\meter}. 
A shorter reactor means a lower conversion but also lower costs. 
The column~(C1) is characterized by the distillate to feed ratio $D/F$ of \num{0.59}.
As a result, \ac{MeOAc} is enriched in the distillate which also contains \ac{MeOH} and \ac{H2O}.
The bottom product is a mixture of \ac{MeOH} and \ac{HOAc}.
In the investigated flowsheet shown in~\autoref{fig:flowsheet_conti}, the bottom product is partially recycled to the feed.
Remarkably, the recycled ratio is set to zero in the depicted best flowsheet. 
These results show that a recycle does not make economic sense for the illustrative flowsheet used for this study.

\begin{table*}
    \centering
    \caption{Continuous design variables selected  by the continuous agent in the best flowsheet observed during \num{10000} episodes of training. }
    \begin{tabular}{l l c c c }
    \toprule
     Unit operation  & Design variable & Symbol & Unit & Best run \\
     \midrule
      Heat exchanger (H1)   & Water inlet temperature & $T^{in}_{\mathrm{water}}$ & \si{\celsius} & 39.7 \\
      Reactor (R1)  & Reactor length & $l$ & \si{\meter} & 5.24\\
      Column (C1)   & Distillate to feed ratio & $D/F$ & - & 0.59 \\
     Recycle & Recycled ratio & - & -&0\\
      \bottomrule
    \end{tabular}
    \label{tab:parameters_conti}
\end{table*}

\subsection{Flowsheet generation in a hybrid action space}
\label{sec:ResultsHybrid}
After the previous sections have shown that all three actors are able to learn separately, they are combined hereinafter.
Therefore, the hybrid agent, combining all previously described elements, is trained in \num{10000} episodes.

\begin{figure}
    \centering
    \includegraphics[width=0.7\textwidth]{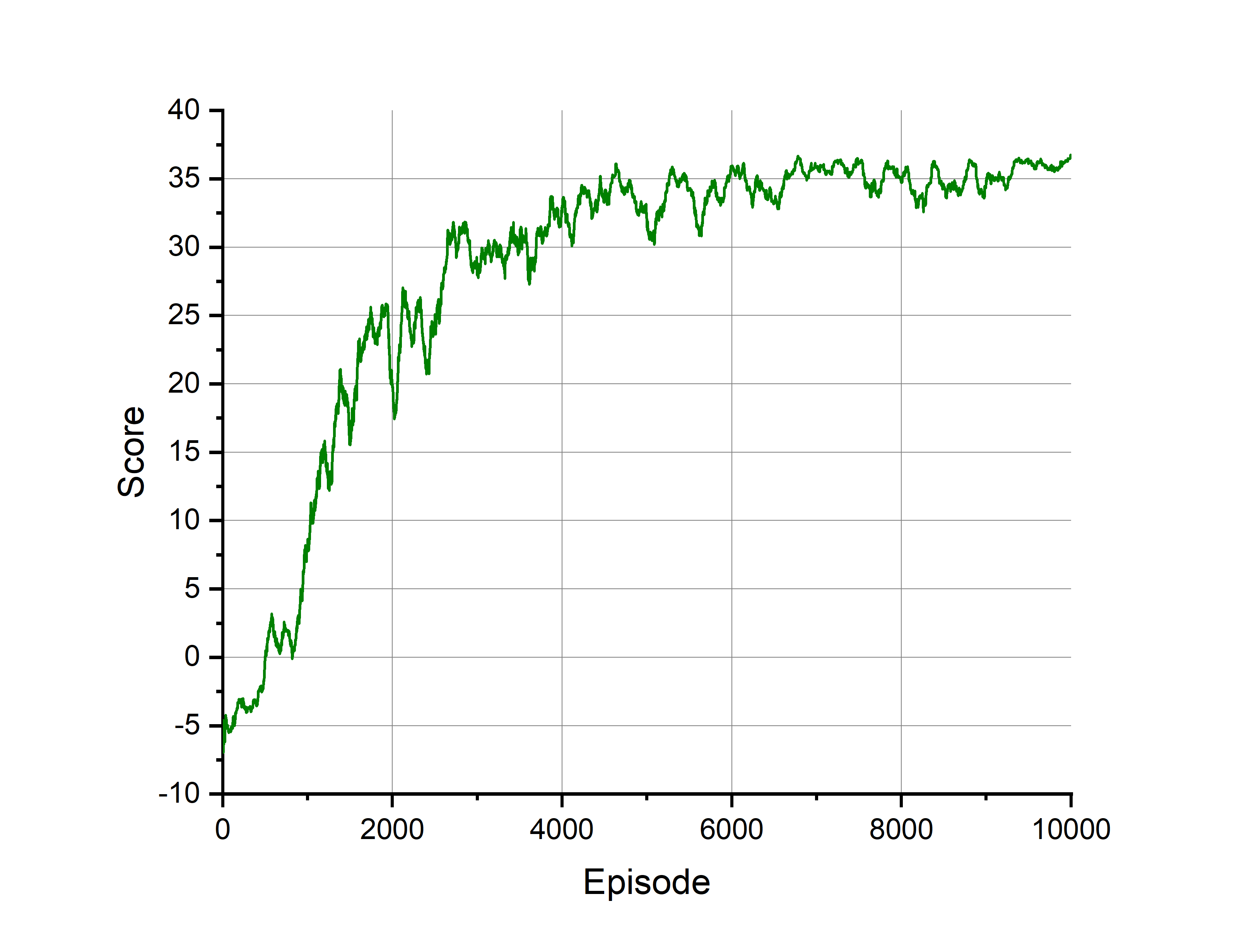}
    \caption{ Learning curve of the agent in a hybrid action space over \num{10000} episodes. Analogously to~\autoref{fig:lc_discrete} and~\autoref{fig:lc_conti}, it shows the scores of the generated flowsheets, averaged over \num{50} episodes.}
    \label{fig:lc_hybrid}
\end{figure}

\begin{figure}
    \centering
    \includegraphics[width=0.35\textwidth]{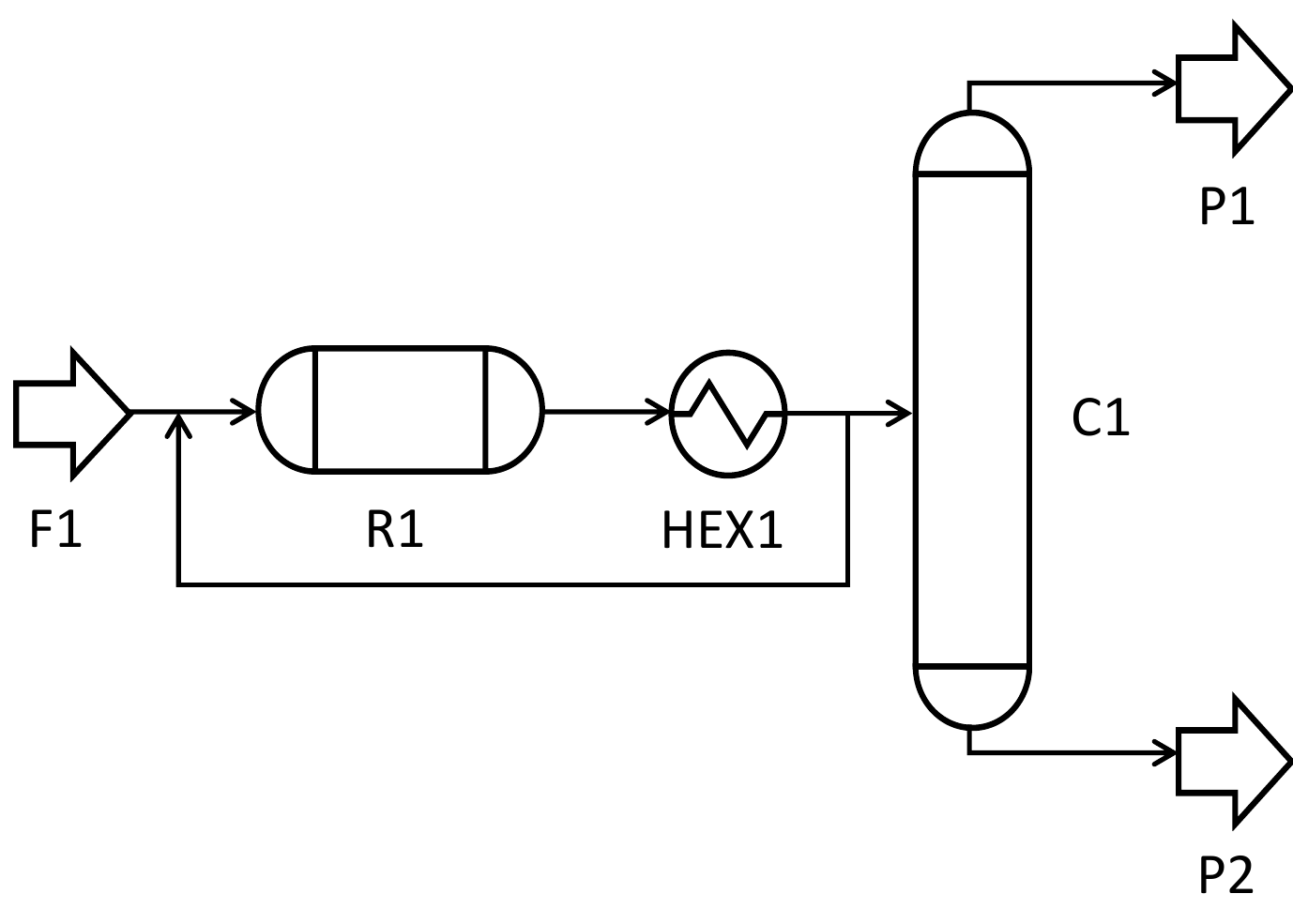}
    \caption{Best flowsheet generated by the agent in a hybrid action space within \num{10000} training episodes. First, \ac{MeOAc} and its side product \ac{H2O} are produced from the feed (F1) in a reactor (R1). Then, the resulting quarternary mixture is heated up in a heat exchanger (HEX1) and split up in a column (C1). Before entering the column, 24\% of the stream are split up and recycled. The first product (P1) is enriched with \ac{MeOAc} but also includes \ac{MeOH} and residues of \ac{H2O}. The second product (P2) is a mixture of \ac{HOAc} and \ac{MeOH}. }
    \label{fig:flowsheets_hybrid}
\end{figure}

The resulting learning curve  is displayed in~\autoref{fig:lc_hybrid}, showing the scores of the flowsheets generated during the training, averaged over \num{50} episodes.
Despite the complexity of the hybrid problem, the agent is learning fast and quickly produces flowsheets with a positive value after approximately \num{1000} episodes.
The best flowsheet the agent observed during training is shown in~\autoref{fig:flowsheets_hybrid}. 
The continuous design variables the agent selected for this best flowsheet are shown in~\autoref{tab:parameters_hybrid}.

The feed (F1) is fed directly into a reactor (R1) where \ac{MeOAc} and \ac{H2O} are produced from esterification of \ac{HOAc} with \ac{MeOH}.
With a length of \SI{18.4}{\meter}, the reactor is significantly larger compared to the best flowsheet generated with the continuous agent in Section~\ref{sec:ResultsConti} which results in a higher conversion but also higher costs.
In the next step, the resulting quarternary mixture is heated in a heat exchanger (HEX1) and split up in a column (C1). 
In the distillate of the column (P1), \ac{MeOAc} is enriched but it also includes \ac{MeOH} and residues of \ac{H2O}.
The bottom product of the column (P2) contains \ac{HOAc} and \ac{MeOH}.
The sequence of unit operations differs from the best flowsheet generated by the discrete agent in Section~\ref{sec:ResultsDiscrete}, were no heat exchanger and two columns were used.
Here, the desired product \ac{MeOAc} is completely in the distillate and the bottom product consists of less valuable chemicals.
Thus, the agent learnt that the second column does not add economic value.
Before entering the column, 24 \% of the process stream are recycled to the feed.
In contrast to the flowsheet investigated in the continuous action space in Section~\ref{sec:ResultsConti}, the recycle does add value to the flowsheet since it increases the total conversion in the reactor.

\begin{table*}
    \centering
    \caption{Continuous design variable selected by the hybrid agent in the best flowsheet observed during \num{10000} episodes of training. }
    \begin{tabular}{l l c c c }
    \toprule
     Unit operation  & Design variable & Symbol & Unit & Best run \\
     \midrule
      Reactor (R1) & Reactor length & $l$ & \si{\meter} & 18.4\\
      Heat exchanger (HEX1)   & Water inlet temperature & $T_{\mathrm{water}}^{in}$ & \si{\celsius} & 36.2\\
      Column (C1)   & Distillate to feed ratio & $D/F$ & - & 0.55\\
      Recycle   & Recycled ratio & - & - & 0.24\\
      \bottomrule
    \end{tabular}
    \label{tab:parameters_hybrid}
\end{table*}

\subsection{Discussion}
Overall, the learning curves shown in the previous sections indicate that all parts of the agent learn quickly. 
It is assumed, however, that the policy does not always converge towards the global optimum for the considered task since the hyperparameters have not been optimized for this first fundamental study.
In future works, it is advised to conduct an extensive hyperparameter study to investigate their influence on the learning behavior.

Compared to other approaches, the main contribution of the presented method is the representation of flowsheets as graphs and combining \acp{GNN} with \ac{RL}.
\acp{GNN} have already shown promising performance in various deep learning tasks~\cite{Zhou.2020}.
One of their key advantage is that they are able to process the topological information of the graphs~\cite{Bronstein.2021}.
Since the structural information about flowsheets is automatically captured in the graph format, \acp{GNN} can take advantage of this structure.
Deriving fingerprints from graphs with \acp{GNN} has already shown promising results in the molecule field~\cite{Duvenaud.2015, Kearnes.2016, Schweidtmann.2020}. 
Here, we transfer the methodology to the flowsheet domain. 
During the implementation and analysis of the training procedure, the graph presentation of the flowsheets has proven to be handy. 
The graphs generated by the agent can be visualized easily and thus immediately give an insight into the process and its meaningfulness. 
An additional advantage of the approach is its flexibility.
Through its hierarchical structure, the different components of the agent can be easily decoupled and new parts can be added.
By using a separate \ac{MLP} for each unit operation in the third level decision, the number of the continuous decisions can vary for the different unit operations.
In the presented work, only one continuous decision is made for each unit operation but the agent architecture allows including more decisions within this step.
By allowing for more unit operations and setting more design variables, the action space and thus the complexity of the problem should be increased for future investigations.

Furthermore, the reward function will require additional attention.
Giving rewards is not straightforward in the considered problem since it is hard to assess the value of an intermediate flowsheet. 
Still, it is crucial for the performance of the \ac{RL} algorithm.
In the presented work, the reward function is only an estimation of economic assessments that neglects multiple cost factors in real processes.
However, for future developments, investigating ways of reward shaping~\cite{Ng.1999} will be an interesting aspect that can stabilize the training process especially when the size of the considered problem gets larger.

\section{Conclusion}
\label{sec:Conclusion}
We propose the first \ac{RL} agent that learns from flowsheet graphs using \acp{GNN} to synthesize new processes. 
The deployed \ac{RL} agent is hierarchical and hybrid meaning it takes multiple dependent discrete and continuous decisions within one step.
In the proposed methodology, the agent first selects a location in an existing flowsheet and a unit operation to extend the flowsheet at the selected position.
Both selections are discrete.
Then, it takes a continuous decision by selecting a design variable that defines the unit operation.
Naturally, each sub-decision strongly depends on the previous one.
Thereby, flowsheets are represented as graphs which allows us to utilize \acp{GNN} within the \ac{RL} structure.
As a result, our methodology generates economical valuable flowsheets only based on experience of the \ac{RL} agent.

In an illustrative case study considering the production of methyl acetate, the approach shows steep and mostly stable learning in discrete, continuous, and hybrid action spaces.
This work is a fundamental study that demonstrates that graph-based \ac{RL} is able to create meaningful flowsheets.
Thus, it encourages to incorporate \ac{AI} in chemical process design.

A further  advantage of the presented approach is that the proposed architecture is a good foundation for further developments like enhancing the state-action space.
Thus, the selected structure of the agent is predestined for increasing the complexity and solving more advanced problems in the future.
A subsequent step following this paper should be to implement an interface to an advanced process simulator.
This will tremendously increase the complexity of the problem but also allow for easier extension of the action space and more rigorous simulations.
As the process simulator will need to deal with random combinations of unit operations, guaranteeing convergence will become a major challenge and including constraints is advisable.

\begin{acknowledgements}
	This work is supported by the TU Delft AI Labs Programme. 
\end{acknowledgements}

\newpage
\section*{Abbreviations}
\begin{acronym}[SMILES]
\acro{AI}[AI]{artificial intelligence}
\acro{ANN}[ANN]{artificial neural network}
\acro{CNN}[CNN]{convolutional neural network}
\acro{GCN}[GCN]{graph convolutional network}
\acro{GCPN}[GCPN]{graph convolutional policy network}
\acro{GNN}[GNN]{graph neural network}
\acro{H2O}[\ce{H2O}]{water}
\acro{HOAc}[HOAc]{acetic acid}
\acro{MDP}[MDP]{Markov decision process}
\acro{MeOAc}[MeOAc]{methyl acetate}
\acro{MeOH}[MeOH]{methanol}
\acro{MINLP}[MINLP]{mixed integer non-linear programming}
\acro{ML}[ML]{machine learning}
\acro{MLP}[MLP]{multi-layer perceptron}
\acro{MPNN}[MPNN]{message passing neural network}
\acro{PFR}[PFR]{plug flow reactor}
\acro{PPO}[PPO]{proximal policy optimization}
\acro{RL}[RL]{reinforcement learning}
\acro{RNN}[RNN]{recurrent neural network}
\acro{SMILES}[SMILES]{simplified molecular-input line-entry system}
\end{acronym}

%
\newpage
\bibliographystyle{Wiley_ama}    
\makeatletter
\renewcommand{\@biblabel}[1]{#1.}
\makeatother
\bibliography{literature}   
\newpage
\section*{Appendix}
\begin{table}[!ht]
    \centering
    \caption{Hyperparameters for the architecture and training procedure of the actor-critic agent.}
    \begin{tabular}{p{10cm}p{2cm}r}
    \toprule
        Parameter &   & Value\\
    \midrule
        Learning rate & $\alpha$ & \num{0.0002}\\
        Policy clipping factor &$\epsilon$ &  \num{0.3}\\
        Discount factor & $\gamma$ & \num{1.0}\\
        $\lambda$-return factor &$\lambda$ &  \num{0.95}\\
        Batch size & $n_{\mathrm{B}}$ & \num{60}\\
        Mini batch size & $n_{\mathrm{MB}}$ & \num{30}\\
        Number of epochs & $n_{\mathrm{E}}$ & \num{4}\\
        
        Weight for loss of level 1 actor & $c_0$ & \num{0.1}\\
        Weight for loss of level 2 actor & $c_1$ &  \num{1.0}\\
        Weight for loss of level 3 actor & $c_2$ & \num{0.5}\\
        Weight for loss of critic & $c_3$ & \num{0.2}\\
        Weight for entropy of level 1 actor & $d_1$ & \num{0.001}\\
        Weight for entropy of level 2 actor & $d_2$ & \num{0.3}\\
        Weight for entropy of level 3 actor & $d_3$ &  \num{0.5}\\
        Weight for entropy of level 3 actor & $d_3$ &  \num{0.001}\\
        
        Hidden layers edge processing for fingerprint & - &  \num{10}\\
        Message passing steps for fingerprint & - &  \num{6}\\
        Hidden layers dimension level 1 actor & - &  \num{12}\\
        Hidden layers dimension level 2 actor & - &  \num{256}\\
        Hidden layers dimension level 3 actor & - &  \num{256}\\
        Hidden layers dimension critic & - &  \num{256}\\
        Feature size flowsheet fingerprint & - &  \num{50}\\
        
    \bottomrule
    \end{tabular}
    \label{tab:hyperparameters}
\end{table}

\end{document}